\documentclass[10pt,twocolumn,letterpaper]{article}

\usepackage{cvpr}              

\usepackage{graphicx}
\usepackage{amsmath}
\usepackage{amssymb}
\usepackage{booktabs}
\usepackage{verbatim}
\usepackage{multirow}
\usepackage{multicol}
\usepackage{bm}

\usepackage{colortbl}
\usepackage{xcolor}
\usepackage{array}

\makeatletter
\robustify\@latex@warning@no@line
\makeatother
\usepackage{authblk}

\newcommand{\legendbox}[1]{%
  \textcolor{#1}{\rule{\fontcharht\font`X}{\fontcharht\font`X}}%
}

\usepackage[pagebackref,breaklinks,colorlinks]{hyperref}

\usepackage[capitalize]{cleveref}
\crefname{section}{Sec.}{Secs.}
\Crefname{section}{Section}{Sections}
\Crefname{table}{Table}{Tables}
\crefname{table}{Tab.}{Tabs.}

\begin{document}

\title{UniVision: A Unified Framework for Vision-Centric 3D Perception}

\author{
    Yu Hong$^{1}$
    \qquad Qian Liu$^{2}$
    \qquad Huayuan Cheng$^1$
    \qquad Danjiao Ma$^2$
    \\Hang Dai$^{3}$
    \qquad Yu Wang$^{2}$
    \qquad Guangzhi Cao$^{2}$
    \qquad Yong Ding$^{1}$
    }

\affil{
    $^{1}$Zhejiang University    
    \qquad $^{2}$Pegasus Tech
    \qquad $^{3}$University of Glasgow
    }


\maketitle

\begin{abstract}
   The past few years have witnessed the rapid development of vision-centric 3D perception in autonomous driving.
   Although the 3D perception models share many structural and conceptual similarities, there still exist gaps in their feature representations, data formats, and objectives, posing challenges for unified and efficient 3D perception framework design.
   In this paper, we present UniVision, a simple and efficient framework that unifies two major tasks in vision-centric 3D perception, \ie, occupancy prediction and object detection. 
   Specifically, we propose an explicit-implicit view transform module for complementary 2D-3D feature transformation.
   We propose a local-global feature extraction and fusion module for efficient and adaptive voxel and BEV feature extraction, enhancement, and interaction.
   Further, we propose a joint occupancy-detection data augmentation strategy and a progressive loss weight adjustment strategy which enables the efficiency and stability of the multi-task framework training.
   We conduct extensive experiments for different perception tasks on four public benchmarks, including nuScenes LiDAR segmentation, nuScenes detection, OpenOccupancy, and Occ3D.
   UniVision achieves state-of-the-art results with +1.5 mIoU, +1.8 NDS, +1.5 mIoU, and +1.8 mIoU gains on each benchmark, respectively.
   We believe that the UniVision framework can serve as a high-performance baseline for the unified vision-centric 3D perception task.
   The code will be available at \url{https://github.com/Cc-Hy/UniVision}.
\end{abstract}

\section{Introduction}

3D perception is the primary task in autonomous driving systems, and its purpose is to use the data obtained by a series of sensors (\eg, LiDAR, Radar, and camera) to derive a comprehensive understanding of the driving scenes, which is used for the subsequent planning and decision-making.
In the past, the field of 3D perception has been dominated by LiDAR-based models due to the accurate 3D information from point cloud data.
However, LiDAR-based systems are costly, vulnerable to bad weather, and inconvenient to deploy.
In comparison, vision-based systems have many advantages such as low cost, easy deployment, and good scalability.
Thus, vision-centric 3D perception has attracted extensive attention from the researchers.

Recently, vision-based 3D detection has been significantly improved via feature representation transformation \cite{cody2021caddn, huang2021bevdet, li2022bevformer}, temporal fusion \cite{Zong2023hop, Wang2023streampetr, Han2023videobev}, and supervision signal design \cite{Hong2022cmkd, monodistill, Chen2022bevdistill}, continuously narrowing the gap with LiDAR-based models.
Lately, vision-based occupancy task has witnessed rapid development \cite{Cao2021monoscene, huang2023tpvformer, wei2023surroundocc, zhang2023occformer, wang2023openoccupancy, tian2023occ3d}. 
Unlike using 3D bounding boxes to represent some whitelist objects, occupancy can more comprehensively describe the geometric and semantics of the driving scene and it is less limited to the shape and category of objects.

Although the detection methods and the occupancy methods share many structural and conceptual similarities, it is not well investigated to simultaneously tackle the two tasks and explore the interrelationship between them.
Occupancy models and detection models often extract different feature representations.
The occupancy prediction task requires exhaustive semantic and geometric judgments across different spatial positions, so the voxel representation is widely used to preserve fine-grained 3D information.
In the detection task, the BEV representation is preferred since most objects are on the same horizontal level with minor overlap.
Compared to the BEV representation, the voxel representation is elaborate but less efficient.
Also, many advanced operators (\eg, shifted window attention \cite{liu2021swin} and deformable convolution \cite{Dai2017dcn}) are primarily designed and optimized for 2D features, making their integration with the 3D voxel representation less straightforward.
The BEV representation is more time-efficient and memory-efficient but it is sub-optimal for dense spatial prediction as it loses structural information in the height dimension.
Apart from feature representations, different perception tasks also differ in their data formats and objectives.
Thus, it is a great challenge to ensure the unity and efficiency of training a multi-task 3D perception framework.

To further exploit the potential of vision models and explore the correlations between different perception tasks, we present UniVision, a unified framework that simultaneously handles the tasks of 3D detection and occupancy prediction. 
The framework follows a join-divide-join diagram. 
Given the surrounding images as input, we use a shared network for image feature extraction.
We propose a novel view transformation module that combines both depth-guided lifting and query-guided sampling for complementary 2D-3D feature transformation.
After that, the network splits into voxel-based and BEV-based branches in parallel to extract features with local and global receptive field awareness, leveraging the advantages of different feature representations.
We then employ adaptive feature interaction between the two feature representations to enhance each other, followed by task-specific heads for different perception tasks.
In addition to the framework design, we present a joint occupancy-detection data augmentation method and a multi-task training strategy for the efficient training of the UniVision framework.
We conduct extensive experiments on four benchmarks, including nuScenes LiDAR segmentation \cite{nuscenes}, nuScenes detection \cite{nuscenes}, OpenOccupancy \cite{wang2023openoccupancy}, and Occ3D \cite{tian2023occ3d}.
The proposed UniVision framework not only efficiently handles different 3D perception tasks, but also achieves state-of-the-art performance on different benchmarks.

Our contributions can be summarized as:
\textbf{i)} 
    We propose a simple and efficient framework for unified vision-centric 3D perception, which simultaneously handles the detection and occupancy tasks.
    Extensive experiments demonstrate the generalization and superiority of the UniVision framework with state-of-the-art performance on different benchmarks.
\textbf{ii)} 
    We propose an explicit-implicit (Ex-Im) view transform method that combines both depth-guided lifting and query-guided sampling, facilitating complementary 2D-3D feature transformation.
\textbf{iii)} 
    We propose a local-global feature extraction and fusion module for efficient and adaptive feature extraction, enhancement, and interaction.
\textbf{iv)}
    We present a joint Occupancy-Detection (Occ-Det) data augmentation method and a progressive loss weight adjustment strategy to enable efficient training of the multi-task framework. 

\section{Related Works}

\subsection{Vision-based 3D Detection}
Vision-based 3D detection aims to locate and classify 3D objects with images from single or multiple cameras.
Early methods \cite{chen2016mono3d, m3drpn, monodle, wang2021fcos3d, monoflex, dd3d} extend advanced 2D object detection methods \cite{tian2019fcos, duan2019centernet} to the 3D case by predicting additional 3D attributes based on 2D ones.
Later, Bird's-Eye-View (BEV) based diagram has become the mainstream.
CaDDN \cite{cody2021caddn} leverages the Lift-Splat-Shoot (LSS) \cite{lss} diagram to transform monocular images into BEV features, executing the detection process within BEV frameworks \cite{pointpillar, second, centerpoint}.
BEVDet \cite{huang2021bevdet} and BEVFormer \cite{li2022bevformer} transform images from surround-view cameras into a single BEV feature map for full-range detection.
Besides, various approaches \cite{zhang2022monodetr, wang2022detr3d, liu2022petr} make efforts to introduce the DETR diagram into the 3D area.
Recent methods have further improved the performance of 3D object detection from the perspective of long-term temporal fusion \cite{Han2023videobev, Zong2023hop, Wang2023streampetr} and sparse representations \cite{Lin2022sparse4d, Liu2023sparsebev}.

\subsection{Vision-based Occupancy Prediction}
Occupancy prediction, also known as semantic scene completion (SSC), requires exhaust judgments for positions in the scene, including whether the position is occupied and the category of occupation.
Early vision-based methods \cite{xiao2010scenessc, song2017semanticssc, gupta2014learningssc, cashman2018contextualssc} use images enriched with additional geometric information, such as RGB-D images, to execute occupancy prediction. 
MonoScene \cite{Cao2021monoscene} is the first method to infer dense geometry and semantics from a single monocular image.
TPVFormer \cite{huang2023tpvformer} enhances the commonly utilized Bird's-Eye-View (BEV) by introducing the Tri-Perspective-View (TPV), thereby augmenting the representation with Z-axis information.
OccFormer \cite{zhang2023occformer} proposes a dual-path transformer network to process the 3D volume for semantic occupancy prediction.
Also, works like OpenOccupancy \cite{wang2023openoccupancy}, Occ3D \cite{tian2023occ3d} and SurroundOcc \cite{wei2023surroundocc} propose pipelines for generating high-quality dense occupancy labels.

\subsection{Multi-task Framework}
Multi-task frameworks \cite{zhang2018multi, kendall2018multi, misra2016crossmulti, ye2022lidarmultinet, chen2018deeplabmulti} strive to efficiently manage various tasks within a singular network.
In the 2D vision area, Mask-RCNN \cite{he_mask_2017} proposes a unified network for object detection and mask segmentation.
UberNet \cite{kokkinos2017ubernetmulti} simultaneously handles a variety of low, medium, and high-level visual tasks in an end-to-end manner.
In LiDAR-based 3D perception, initiatives such as LidarMTL \cite{Feng2021lidarmtl} and LidarMultiNet \cite{ye2022lidarmultinet} leverage a shared network for tasks encompassing 3D detection, segmentation, and road understanding.
A major advantage of multi-task networks is to save computational and storage overheads by utilizing shared model structure and weights.
However, the performance of individual tasks frequently diminishes as the network navigates trade-offs between different objectives, posing challenges for multi-task framework design.


\begin{figure*}
    \centering
    \includegraphics[width=1\linewidth]{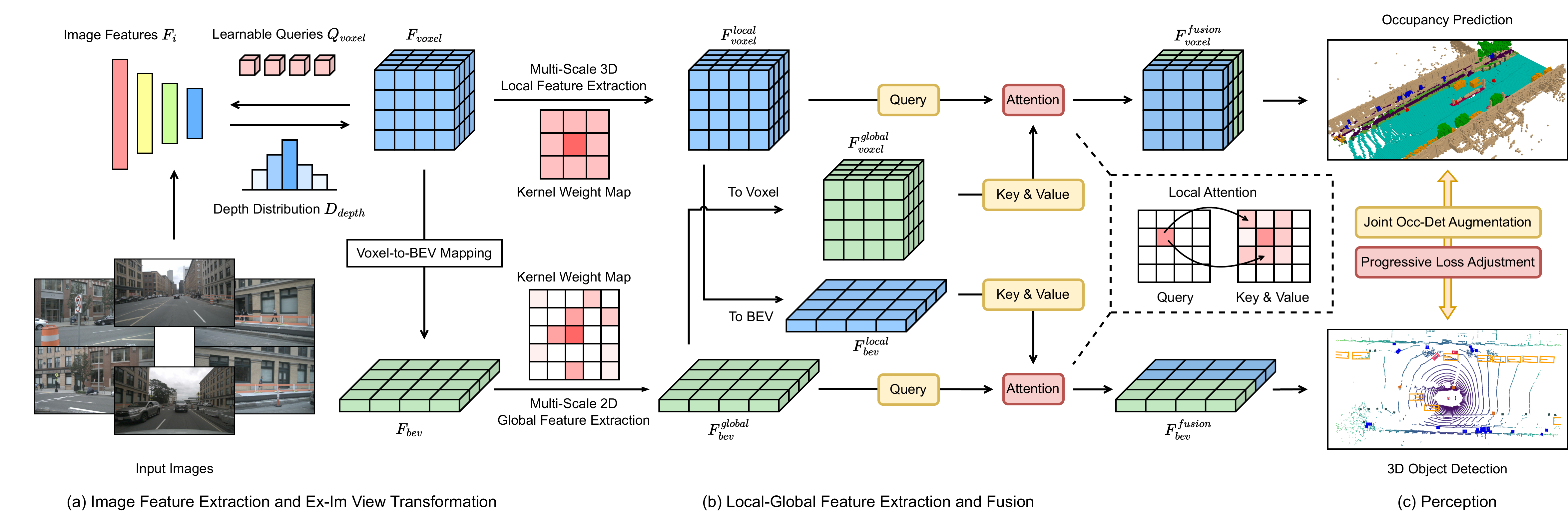}
    \caption{
    The overall architecture of UniVision. 
    After extracting image features from the inputs, we use the Ex-Im view transform module for complementary 2D-3D feature transformation.
    We then propose the local-global feature extraction and fusion block for adaptive BEV and voxel feature extraction, enhancement, and interaction, which are attached to task-specific perception heads.
    During training, the joint Occ-Det augmentation and progressive loss weight adjustment strategy are equipped for efficient multi-task training.
    }
    \label{fig:overall}
\end{figure*}


\section{Method}

\subsection{Framework Overview}
\cref{fig:overall} shows the overall architecture of the UniVision framework.
Given multi-view images $\{I^i | I^i \in\mathbb{R}^{W_I\times H_I\times3\}}$, $i\in[1,N]$ from the surrounding $N$ cameras as input, an image feature extraction network is first utilized to extract image features $F_{img}$ from them.
The 2D image features $F_{img}$ are then lifted to 3D voxel features $F_{voxel}$ with the Ex-Im view transform module, which combines depth-guided explicit feature lifting and query-guided implicit feature sampling.
The voxel features $F_{voxel}$ are sent into the local-global feature extraction and fusion block to extract local-context-aware voxel features and global-context-aware BEV features respectively.
Then, we employ the cross-representation feature interaction module to perform information exchange on the voxel features and BEV features, which are used for different downstream perception tasks.
During training, the joint Occ-Det data augmentation and the progressive loss weight adjustment strategy are used for efficient training of the UniVision framework.

\begin{figure*}
    \centering
    \includegraphics[width=1\linewidth]{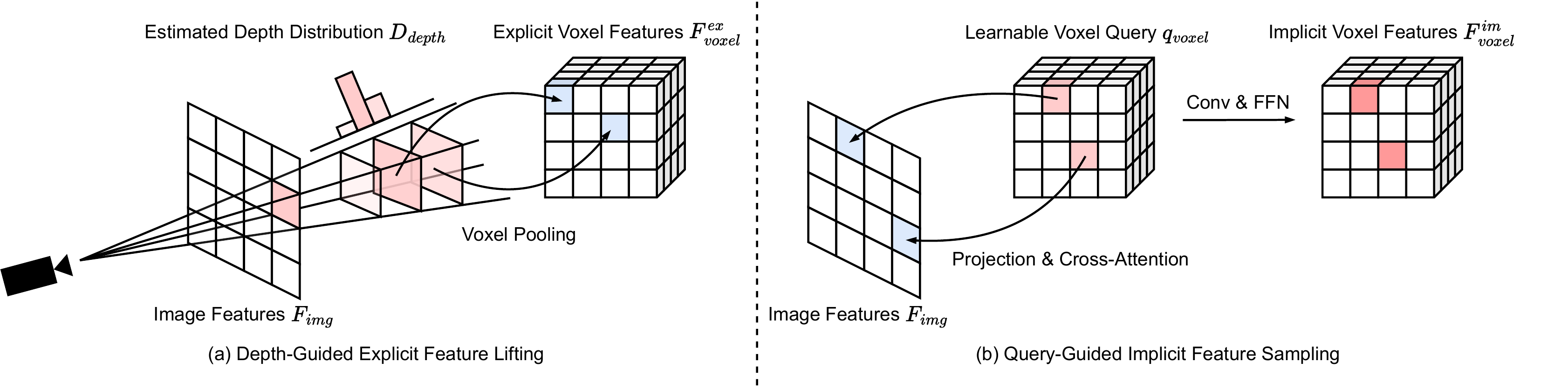}
    \caption{The Ex-Im view transform module. (a) Depth-guided Explicit Feature Lifting. (b) Query-guided Implicit Feature Sampling.}
    \label{fig:view-transform}
\end{figure*}

\subsection{Ex-Im View Transform}

\textbf{Depth-guided Explicit Feature Lifting.}
Following the Lift-Splat-Shoot (LSS) \cite{lss} diagram, 
we perform the voxel pooling operation \cite{huang2021bevdet} on the per-pixel depth distribution $D_{depth} \in \mathbb{R}^{D\times H\times W}$ and the image features $F_{img} \in \mathbb{R}^{C\times H\times W}$ to extract the voxel features:
\begin{equation}
    F_{voxel}^{ex} = VoxelPooling(D_{depth}, F_{img})
\end{equation}
Since $F_{voxel}^{ex}$ is generated with the explicit depth distribution estimation, we refer to it as the explicit voxel features.

\textbf{Query-guided Implicit Feature Sampling.}
However, $F_{voxel}^{ex}$ has some defects in representing the 3D information.
The accuracy of $F_{voxel}^{ex}$ is highly related to the accuracy of the estimated depth distribution $D_{depth}$.
Also, the points generated from LSS are unevenly distributed. 
Points are dense close to the camera and are sparse in distance.
We thus further use the query-guided feature sampling to compensate for the above shortcomings of $F_{voxel}^{ex}$.
We define the learnable voxel queries $q_{voxel} \in \mathbb{R}^{C\times X\times Y \times Z}$, and use a 3D transformer to sample the features from the images.
For each voxel query, we project its center $\mathbf{c}$ onto the image plane with calibration matrix $\mathbf{P}$ for the reference point $\mathbf{p}$, and then use $N$ transformer blocks. Each block includes a deformable cross-attention (DCA) \cite{zhu2020deformabledetr} layer, a 3D convolution (Conv) layer, and a feed-forward network (FFN) \cite{vaswani2017attention}:
\begin{gather}
    \mathbf{p} = \mathbf{P\times {c}} \\
    q^{i+1} = FFN(Conv(DCA(q^i, \mathbf{p}, F_{img}))) \\
    F_{voxel}^{im} = q^N
\end{gather}
Compared to the points generated from LSS, the voxel queries are evenly distributed in the 3D space and they are learned from the statistical properties of all training samples, which is independent of the depth prior information used in LSS.
Thus, $F_{voxel}^{ex}$ and $F_{voxel}^{im}$ complement each other, and we concatenate them as the output features of the view transform module:
\begin{equation}
    F_{voxel} = F_{voxel}^{ex}\mathbin\Vert F_{voxel}^{im}
\end{equation}
where $\mathbin\Vert$ is the concatenate operation. 
The Ex-Im view transform module is illustrated in \cref{fig:view-transform}.


\subsection{Local-Global Feature Extraction and Fusion}
Given input voxel features $F_{voxel} \in \mathbb{R}^{C\times X\times Y \times Z}$, we first stack the features in Z axis and use a convolution layer to reduce the channels to obtain the BEV features $F_{bev} \in \mathbb{R}^{C\times X\times Y}$:
\begin{gather}
    \begin{aligned}
    F_{bev} &= Conv(Stack(F_{voxel}, dim=Z))
    \end{aligned}
\end{gather}
Then, the model splits into two parallel branches for feature extraction and enhancement.

\textbf{Local feature extraction.} 
For $F_{voxel}$, we use a local feature extraction branch composed of 3D convolutions to extract local features of each spatial position.
We extend ResNet \cite{resnet} to ResNet3D to extract multi-scale voxel features $\{F_{voxel}^i \,|\, F_{voxel}^i \in \mathbb{R}^{C*2^i\times \frac{X}{2^i} \times \frac{Y}{2^i} \times \frac{Z}{2^i}}\}$ from $F_{voxel}$.
We then use the FPN \cite{fpn} structure from SECOND \cite{second} in 3D version to merge $\{F_{voxel}^i\}$ into $F_{voxel}^{local} \in \mathbb{R}^{C\times X\times Y \times Z}$.

\textbf{Global feature extraction.}
The BEV features $F_{bev}$ retain the information at the object level and they are computationally efficient. 
Thus, We propose a global feature extraction branch to extract features with a global receptive field based on the BEV representation.
We use a network with deformable convolution v3 (DCNv3) \cite{wang2022internimage} to dynamically gather global information for multi-scale BEV features $\{F_{bev}^i \,|\, F_{bev}^i \in \mathbb{R}^{C*2^i\times \frac{X}{2^i} \times \frac{Y}{2^i}}\}$.
And $\{F_{bev}^i\}$ goes through the SECOND FPN structure for the merged BEV feature $F_{bev}^{global} \in \mathbb{R}^{C\times X\times Y }$.

\textbf{Cross-Representation Feature Interaction.}
After generating the local-context-aware voxel features $F_{voxel}^{local}$ and the global-context-aware BEV features $F_{bev}^{global}$ from the input voxel features, 
we use the cross-representation feature interaction module to enable adaptive information exchange between the two feature representations for further enhancement.
We first map the BEV features $F_{bev}^{global}$ to voxel features $F_{voxel}^{global}$ and map voxel features $F_{voxel}^{local}$ to BEV features $F_{voxel}^{local}$ by Z-dimensional repetition or addition:
\begin{align}
    F_{voxel}^{global} &= repeat(F_{bev}^{global}, dim=Z)                    \\
    F_{bev}^{local} &= add(F_{voxel}^{local}, dim=Z)
\end{align}
For the voxel representation, we use $F_{voxel}^{local}$ from the voxel branch as the query, and $F_{voxel}^{global}$ from the BEV branch as the key and value.
And we generalize the neighborhood attention transformer \cite{hassani2023neighborhoodattention} from self-attention to cross-attention to perform information gathering within a local perception field $\Delta \bm p$, and a symmetric process is applied on the BEV features:
\begin{align}
    F_{voxel}^{fusion}  = Attn(q=F_{voxel}^{local}, k\&v=F_{voxel}^{global}, \Delta \bm{p}) \\
    F_{bev}^{fusion}  = Attn(q=F_{bev}^{global}, k\&v=F_{bev}^{local}, \Delta \bm{p})
\end{align}
Specifically, we set $\Delta \bm p$ to $3\times 3$ for the BEV features and  $3\times 3 \times 3$ for the voxel features.

\subsection{Heads and Losses}
We attach task-specific heads to $F_{voxel}^{fusion}$ and $F_{bev}^{fusion}$ for different perception tasks. 
For the occupancy task, we use two fully connected layers to map the feature channels to the number of occupancy categories.
We follow the loss setting in OpenOccupancy \cite{wang2023openoccupancy}, which combines the cross-entropy loss, Lovasz softmax loss, geometry affinity loss, and semantic affinity loss:
\begin{equation}
    \mathcal{L}_{occ} = \lambda_1\cdot L_{ce} + \lambda_2\cdot L_{lovasz} + \lambda_3\cdot L_{geo} + \lambda_4\cdot L_{sem}
\end{equation}
For the detection task, we use the center-based head \cite{centerpoint} and the detection loss is composed of the classification loss $L_{cls}$ and the regression loss $L_{reg}$:
\begin{equation}
    \mathcal{L}_{det} = \lambda_5\cdot L_{cls} + \lambda_6\cdot L_{reg}
\end{equation}
Also, we add the depth loss $L_{depth}$ used in BEVDepth \cite{li2022bevdepth} as the image level supervision:
\begin{equation}
    \mathcal{L}_{img} = \lambda_7\cdot L_{depth}
\end{equation}
The overall loss function is defined as:
\begin{equation}
    \mathcal{L} = \mathcal{L}_{img} + \mathcal{L}_{det} + \mathcal{L}_{occ}
\end{equation}

\begin{figure*}
    \centering
    \includegraphics[width=1\textwidth]{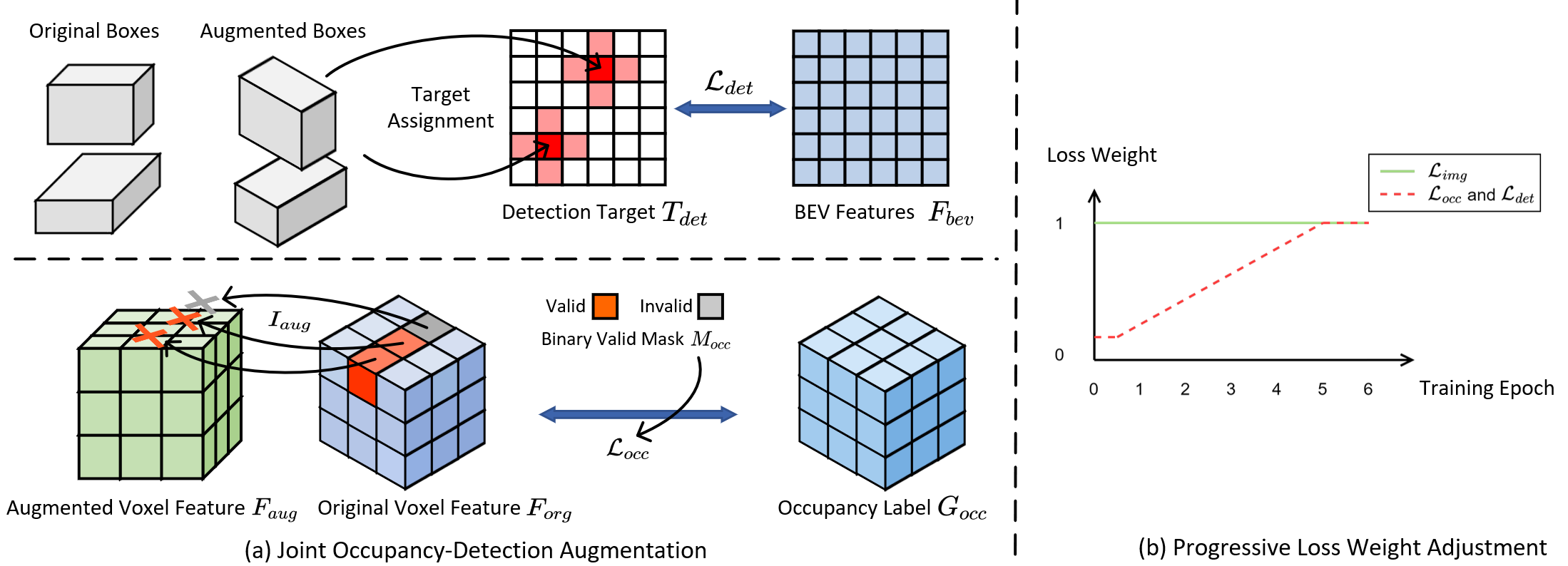}
    \caption{Illustration of (a) joint occupancy-detection augmentation and (b) progressive loss weight adjustment strategy.}
    \label{fig:occ-det aug and progressive-loss-weight}
\end{figure*}

\textbf{Progressive Loss Weight Adjustment Strategy.} 
In practice, we find that directly combining the above losses tends to lead to a failed training process and the network cannot converge.
In the early stage of training, the voxel features $F_{voxel}$ are randomly distributed, and the supervision in the occupancy head and detection head contribute less than other losses in the convergence. 
Meanwhile, the loss items like the classification loss $L_{cls}$ in the detection task
are very large and dominate the training process, making the model difficult to optimize.

To overcome this, we propose the progressive loss weight adjustment strategy to dynamically adjust the loss weights.
Specifically, a control parameter $\delta$ is added to the non-image-level losses, \ie, the occupancy loss and the detection loss, to adjust loss weights in different training periods.
The control weight $\delta$ is set to a small value $V_{min}$ at the beginning and gradually increase to $V_{max}$ in $N$ training epoches:
\begin{gather}
    \mathcal{L} = \mathcal{L}_{img} + \delta\cdot\mathcal{L}_{det} + \delta\cdot\mathcal{L}_{occ} \\
    \delta = max(V_{min}, min(V_{max}, \frac{i}{N}\cdot V_{max}))
\end{gather}
where $i$ denotes the $i_{th}$ training epoch.
In this case, the optimization process first focuses on the image-level information (depth and semantics) to generate reasonable voxel representations, and then on the subsequent perception tasks.
The progression is illustrated in \cref{fig:occ-det aug and progressive-loss-weight} (b).

\subsection{Joint Occ-Det Spatial Data Augmentation}
In the 3D detection task, spatial-level data augmentation is also effective in improving model performance in addition to the common image-level data augmentation.
However, it is not straightforward to apply spatial-level augmentation in the occupancy task.
When we apply data augmentation such as random scaling and rotation to the discrete occupancy labels $G_{occ} \in \mathbb{R}^{X\times Y\times Z}$, it is difficult to determine the semantics of the generated voxels.
Thus, the existing methods only apply simple spatial augmentation like random flipping in occupancy tasks.

To address this problem, we propose a joint Occ-Det spatial data augmentation to allow simultaneous augmentation in both the 3D detection task and the occupancy task in our UniVision framework.
Since the 3D box labels are in continuous values and the augmented 3D box can be directly calculated for training, we follow the augmentation method for detection from BEVDet \cite{huang2021bevdet}.
Although the occupancy labels are discrete and difficult to operate on, the voxel features can be regarded as continuous and can be handled with operations like sampling and interpolation.
Thus, we propose to transform the voxel features instead of directly operating on the occupancy labels for the data augmentation.

Specifically, we first sample spatial data augmentations and calculate the corresponding 3D transformation matrix $\mathbf{M}_{aug}$.
For the occupancy labels $G_{occ} \in \mathbb{R}^{X\times Y\times Z}$ and its voxel indices $\mathbf{I}_{org} \in \mathbb{R}^{X\times Y\times Z\times 3}$, we compute the their 3D coordinates $\mathbf{C}_{org} \in \mathbb{R}^{X\times Y\times Z\times 3}$.
We then apply $\mathbf{M}_{aug}$ to $\mathbf{C}_{org}$, and normalize them to obtain the voxel indices $\mathbf{I}_{aug}$ in the augmented voxel features:
\begin{gather}
    \mathbf{C}_{org} = \mathbf{P}_{i-c}\times \mathbf{I}_{org} \\
    \mathbf{I}_{aug} = \mathbf{P}_{c-i}\times \mathbf{M}_{aug}\times \mathbf{C}_{org}
\end{gather}
where $\mathbf{P}_{i-c}$ and $\mathbf{P}_{c-i}$ are the transformation matrices between voxel indices and spatial coordinates.
Then, we sample the voxel features $F_{aug}$ with the voxel indices $\mathbf{I}_{aug}$:
\begin{gather}
    F_{org} = S(F_{aug}, \mathbf{I}_{org})
\end{gather}
where $S$ denotes sampling operation and $F_{org}$ are the sampled voxel features that correspond to the original occupancy labels $G_{occ}$ without transformation, which can be used for loss calculation.
Noticeably, some sampling positions can fall out of range, and we ignore these voxels by adding a binary mask $M_{occ}\in \{0,1\}^{X\times Y\times Z}$ when calculating the occupancy losses:
\begin{equation}
    \mathcal{L}_{occ} = f(G_{occ}, F_{org}) \times M_{occ}
\end{equation}
where $f$ denotes the loss functions. 
We illustrate the joint Occ-Det augmentation in \cref{fig:occ-det aug and progressive-loss-weight} (a).

\begin{table*}[ht]
    \renewcommand{\arraystretch}{0.8}
    \footnotesize
    \centering
    \setlength{\tabcolsep}{1.5mm}{
    \resizebox{\linewidth}{!}{
    \begin{tabular}{l|c|c|cccccccccccccccc}
    \toprule
    Method & Modality & mIoU 
    & $\rotatebox{90}{\legendbox{orange} barrier}$ 
    & $\rotatebox{90}{\legendbox{pink} bycicle}$ 
    & $\rotatebox{90}{\legendbox{yellow} bus}$ 
    & $\rotatebox{90}{\legendbox{blue} car}$ 
    & $\rotatebox{90}{\legendbox{cyan} cons. vehi.}$ 
    & $\rotatebox{90}{\legendbox{magenta} motorcycle}$ 
    & $\rotatebox{90}{\legendbox{black} pedestrian}$ 
    & $\rotatebox{90}{\legendbox{red} traffic\_cone}$ 
    & $\rotatebox{90}{\legendbox{brown} trailer}$
    & $\rotatebox{90}{\legendbox{purple} truck}$ 
    & $\rotatebox{90}{\legendbox{pink} driv. surf.}$ 
    & $\rotatebox{90}{\legendbox{violet} other flat}$ 
    & $\rotatebox{90}{\legendbox{teal} sidewalk}$ 
    & $\rotatebox{90}{\legendbox{olive} terrain}$ 
    & $\rotatebox{90}{\legendbox{lightgray} manmade}$ 
    & $\rotatebox{90}{\legendbox{green} vegetation}$ 
    \\
    \midrule
    PolarNet \cite{zhang2020polarnet} &LiDAR &69.4 &72.2 &16.8 &77.0 &86.5 &51.1 &69.7 &64.8 &54.1 &69.7 &63.5 &96.6 &67.1 &77.7 &72.1 &87.1 &84.5\\
    DB-UNet \cite{wang2023dbunet} &LIDAR & 70.1 & 67.5 &23.7 &75.3 &82.1 &47.0 &72.5 &67.3 &66.6 &74.3 &60.1 &96.9 &64.2 &76.9 &73.4 &88.0 &86.7\\
    PolarStream \cite{zhang2020polarnet} &LiDAR &73.4 &71.4 &27.8 &78.1 &82.0 &61.3 &77.8 &75.1 &72.4 &79.6 &63.7 &96.0 &66.5 &76.9 &73.0 &88.5 &84.8\\
    Cylinder3D++ \cite{zhou2020cylinder3d}  &LiDAR   &77.9 &82.8 &33.9 &84.3 &89.4 &69.6 &79.4 &77.3 &73.4 &84.6 &69.4 &97.7 &70.2 &80.3 &75.5 &90.4 &87.6\\
    LidarMultiNet \cite{ye2022lidarmultinet} & LiDAR &81.4 &80.4 &48.4 &94.3 &90.0 &71.5 &87.2 &85.2 &80.4 &86.9 &74.8 &97.8 &67.3 &80.7 &76.5 &92.1 &89.6\\
    \midrule
    TPVFormer \cite{huang2023tpvformer}  &Camera &69.4 &\textbf{74.0} &27.5 &86.3 &85.5& \textbf{60.7} &68.0 &62.1 &49.1 &81.9 &68.4 &94.1 &59.5 &66.5 &63.5 &83.8 &79.9\\
    OccFormer \cite{zhang2023occformer}  &Camera   &70.8     &72.8 &29.9 &\textbf{87.9} &85.6& 57.1 &74.9 &63.2 &53.5 &83.0 &67.6 &94.8 &61.9 &70.0 &\textbf{66.0} &\textbf{84.0} &\textbf{80.5}\\
    \rowcolor{gray!20} 
    UniVision  &Camera &\textbf{72.3}     &72.1 &\textbf{34.0} &85.5 &\textbf{89.5} & 59.3 &\textbf{75.5} &\textbf{69.3} &\textbf{65.8} &\textbf{84.2} &\textbf{71.4} &\textbf{96.1} &\textbf{67.4} &\textbf{71.9} &65.0 &77.9 &71.7\\
    \bottomrule
    \end{tabular}}}
    \caption{Results on nuScenes LiDAR segmentation benchmark (test). 
    UniVision sets a new record on the leaderboard
    among camera-based methods and has comparable or better results than some of the LiDAR-based methods.
    }
    \label{tab:benchmark-nus-lidarseg}
\end{table*}

\begin{table*}[ht]
    \renewcommand{\arraystretch}{0.8}
    \footnotesize
    \centering
    \begin{tabular}{l|c|c|cc|ccccc}
    \toprule
    Method & Backbone & Resolution &mAP$\uparrow$ &NDS$\uparrow$ &mATE$\downarrow$ &mASE$\downarrow$ &mAOE$\downarrow$ &mAVE$\downarrow$ &mAAE$\downarrow$ \\
    \midrule
    BEVFormer* \cite{li2022bevformer} &R50  &$256\times 704$      &0.232 &0.339 &0.852 &0.308 &0.720 &\textbf{0.648} &0.240 \\
    BEVDet* \cite{huang2021bevdet}    &R50   &$256\times 704$      &0.278 &0.348 &0.783 &0.289 &0.686 &0.860 &0.289 \\
    BEVDepth* \cite{li2022bevdepth}   &R50  &$256\times 704$       &0.292 &0.386 &0.724 &\textbf{0.269} &\textbf{0.575} &0.796 &0.241 \\
    \rowcolor{gray!20} 
    UniVision*                        &R50 &$256\times 704$          &\textbf{0.312} &\textbf{0.397} &\textbf{0.682} &0.287 &0.632 &0.763 &\textbf{0.224} \\
    \midrule
    BEVFormer* \cite{li2022bevformer} &R50  &$512\times 1408$     &0.292 &0.388 &0.819 &0.294 &0.635 &\textbf{0.605} &\textbf{0.222} \\
    BEVDet* \cite{huang2021bevdet}    &R50  &$512\times 1408$        &0.329 &0.394 &0.724 &0.275 &0.607 &0.852 &0.247 \\
    BEVDepth* \cite{li2022bevdepth}   &R50   &$512\times 1408$      &0.354 &0.428 &0.674 &\textbf{0.267} &\textbf{0.506} &0.806 &0.236 \\
    \rowcolor{gray!20} 
    UniVision*                        &R50  &$512\times 1408$     &\textbf{0.378} &\textbf{0.439} &\textbf{0.658} &0.273 &0.559 &0.748 &0.260 \\
    \midrule
    PolarFormer  \cite{jiang2022polar} &R101  &$900\times 1600$ &0.396 &0.458 &0.700 &0.269 &0.375 &0.839 &0.245 \\
    FCOS3D     \cite{wang2021fcos3d}                   &R101  &$900\times 1600$     &0.343 &0.415 &0.725 &0.263 &0.422 &1.292 &\textbf{0.153} \\
    PGD      \cite{wang2022pgd}                     &R101  &$900\times 1600$     &0.369 &0.428 &0.683 &\textbf{0.260} &0.439 &1.268 &0.185 \\
    DETR3D\cite{wang2022detr3d}                           &R101  &$900\times 1600$     &0.346 &0.425 &0.773 &0.268 &0.383 &0.842 &0.216 \\
    PETR \cite{liu2022petr}                       &R101  &$512\times 1408$     &0.357 &0.421 &0.710 &0.270 &0.490 &0.885 &0.224 \\
    BEVDet \cite{huang2021bevdet}  &SwinT &$512\times1408$ &0.349 &0.417 &0.637 &0.269 &0.490 &0.914 &0.268\\
    BEVDepth \cite{li2022bevdepth} &R101 &$512\times1408$ &0.376 &0.408 &0.659 &0.267 &0.543 &1.059 &0.335 \\
    \rowcolor{gray!20}
    UniVision                        &R101  &$512\times 1408$  &\textbf{0.413} &\textbf{0.490} &\textbf{0.600} &0.263 &\textbf{0.366} &\textbf{0.731} &0.211   \\
    \midrule
    PETRv2 \cite{liu2022petrv2}  &R101  &$640\times1600$ &0.421 &0.524 &0.681 &0.267 &0.357 &\textbf{0.377} &\textbf{0.186} \\
    PolarFormer   \cite{jiang2022polar} &R101 &$900\times1600$  &0.432 &0.528 &0.648 &0.270 &0.348 &0.409 &0.201 \\
    BEVFormer \cite{li2022bevformer}  &R101 &$900\times1600$ &0.416 &0.517 &0.673 &0.274 &0.372 &0.394 &0.198 \\
    BEVDet4D \cite{huang2022bevdet4d}  &Swin-B &$640\times1600$ &0.396 &0.515 &0.619 &\textbf{0.260} &0.361 &0.399 &0.189 \\
    \rowcolor{gray!20}
    UniVision4D (2frame)                 &R101  &$512\times 1408$  &0.439 &0.535 &0.580 &0.264 &\textbf{0.310} &0.491 &0.200   \\
    \rowcolor{gray!20}
    UniVision4D (4frame)                &R101  &$512\times 1408$   &\textbf{0.452} &\textbf{0.546} &\textbf{0.556} &0.264 &0.330 &0.451 &0.199  \\
    \bottomrule
    \end{tabular}
    \caption{
    Results on nuScenes detection benchmark (val). 
    Methods with * are trained with aligned training settings, including input resolution, backbone, batch size, learning rate, etc. for a fair comparison.
    }
    \label{tab:benchmark-nus-det}
\end{table*}

\section{Experiments and Discussions}
\subsection{Datasets and Evaluation Metrics}
\textbf{NuScenes.} 
NuScenes \cite{nuscenes} is a modern and large-scale dataset for autonomous driving which contains 1000 driving scenes.
It provides sensor data including LiDAR, radar, camera, and support benchmarking for different autonomous driving tasks, \eg, 3D detection, LiDAR segmentation, and motion planning.

\textbf{NuScenes LiDAR Segmentation.}
Following recent OccFormer \cite{zhang2023occformer} and TPVFormer \cite{huang2023tpvformer}, we use camera images as input for the LiDAR segmentation task, and the LiDAR data is only used to provide 3D positions for querying the output features.
We use the mean intersection over union (mIoU) as the evaluation metric.

\begin{table*}[ht]
    \renewcommand{\arraystretch}{0.8}
    \footnotesize
    \centering
    \setlength{\tabcolsep}{1.5mm}{
    \resizebox{\linewidth}{!}{
    \begin{tabular}{l|c|c|cccccccccccccccc}
    \toprule
    Method & Modality & mIoU
    & $\rotatebox{90}{\legendbox{orange} barrier}$ 
    & $\rotatebox{90}{\legendbox{pink} bycicle}$ 
    & $\rotatebox{90}{\legendbox{yellow} bus}$ 
    & $\rotatebox{90}{\legendbox{blue} car}$ 
    & $\rotatebox{90}{\legendbox{cyan} cons. vehi.}$ 
    & $\rotatebox{90}{\legendbox{magenta} motorcycle}$ 
    & $\rotatebox{90}{\legendbox{black} pedestrian}$ 
    & $\rotatebox{90}{\legendbox{red} traffic\_cone}$ 
    & $\rotatebox{90}{\legendbox{brown} trailer}$ 
    & $\rotatebox{90}{\legendbox{purple} truck}$ 
    & $\rotatebox{90}{\legendbox{pink} driv. surf.}$ 
    & $\rotatebox{90}{\legendbox{violet} other flat}$ 
    & $\rotatebox{90}{\legendbox{teal} sidewalk}$ 
    & $\rotatebox{90}{\legendbox{olive} terrain}$ 
    & $\rotatebox{90}{\legendbox{lightgray} manmade}$ 
    & $\rotatebox{90}{\legendbox{green} vegetation}$ 
    \\
    \midrule
    LMSCNet \cite{roldao2020lmscnet} &LiDAR &11.5 &12.4 &4.2 &12.8 &12.1 &6.2 &4.7 &6.2 &6.3 &8.8 &7.2 &24.2 &12.3 &16.6 &14.1 &13.9 &22.2 \\
    JS3C-Net \cite{yan2021js3cnet} &LIDAR &12.5 &14.2 &3.4 &13.6 &12.0 &7.2 &4.3 &7.3 &6.8 &9.2 &9.1 &27.9 &15.3 &14.9 &16.2 &14.0 &24.9 \\
    L-CONet \cite{wang2023openoccupancy} &LiDAR &15.8 &17.5 &5.2 &13.3 &18.1 &7.8 &5.4 &9.6 &5.6 &13.2 &13.6 &34.9 &21.5 &22.4 &21.7 &19.2 &23.5\\
    \midrule
    AICNet \cite{li2020aicnet} &Camera \& Depth   &10.6 &11.5 &4.0 &11.8 &12.3 &5.1 &3.8 &6.2 &6.0 &8.2 &7.5 &24.1 &13.0 &12.8 &11.5 &11.6 &20.2  \\   
    3DSketch \cite{chen20203dsketch} &Camera \& Depth &10.7 &12.0 &5.1 &10.7 &12.4 &6.5 &4.0 &5.0 &6.3 &8.0 &7.2 &21.8 &14.8 &13.0 &11.8 &12.0 &21.2  \\
    \midrule
    MonoScene \cite{Cao2021monoscene}  &Camera  &6.9 &7.1 &3.9 &9.3 &7.2 &5.6 &3.0 &5.9 &4.4 &4.9 &4.2 &14.9 &6.3 &7.9 &7.4 &\textbf{10.0} &7.6\\
    TPVFormer \cite{huang2023tpvformer}  &Camera &7.8 &9.3 &4.1 &11.3 &10.1 &5.2 &4.3 &5.9 &5.3 &\textbf{6.8} &6.5 &13.6 &9.0 &8.3 &8.0 &9.2 &8.2\\
    C-CONet \cite{wang2023openoccupancy} &Camera &12.8 &13.2 &8.1 &\textbf{15.4} &17.2 &6.3 &11.2 &10.0 &8.3 &4.7 &12.1 &\textbf{31.4} &18.8 &18.7 &16.3 &4.8 &8.2 \\
    \rowcolor{gray!20}
    UniVision  &Camera &\textbf{14.3}     &\textbf{14.7} &\textbf{9.6} &12.9 &\textbf{17.2} & \textbf{8.6} &\textbf{12.1} &\textbf{12.1} &\textbf{9.1} &6.3 &\textbf{13.0} &30.5 &\textbf{23.0} &\textbf{19.9} &\textbf{18.0} &9.1 &\textbf{13.3}\\
    \bottomrule
    \end{tabular}}}
    \caption{Results on OpenOccupancy benchmark. UniVision achieves state-of-the-art performance among camera-based methods and has comparable or better results than some of the LiDAR-based methods.
    }
    \label{tab:benchmark-nus-openoccupancy}
\end{table*}

\begin{table*}[ht]
    \renewcommand{\arraystretch}{0.8}
    \footnotesize
    \centering
    \setlength{\tabcolsep}{1.5mm}{
    \resizebox{\linewidth}{!}{
    \begin{tabular}{l|c|c|ccccccccccccccccc}
    \toprule
    Method & Resolution & mIoU
    & $\rotatebox{90}{\legendbox{black} others}$ 
    & $\rotatebox{90}{\legendbox{orange} barrier}$ 
    & $\rotatebox{90}{\legendbox{pink} bycicle}$ 
    & $\rotatebox{90}{\legendbox{yellow} bus}$ 
    & $\rotatebox{90}{\legendbox{blue} car}$ 
    & $\rotatebox{90}{\legendbox{cyan} cons. vehi.}$ 
    & $\rotatebox{90}{\legendbox{magenta} motorcycle}$ 
    & $\rotatebox{90}{\legendbox{black} pedestrian}$ 
    & $\rotatebox{90}{\legendbox{red} traffic\_cone}$ 
    & $\rotatebox{90}{\legendbox{brown} trailer}$
    & $\rotatebox{90}{\legendbox{purple} truck}$ 
    & $\rotatebox{90}{\legendbox{pink} driv. surf.}$ 
    & $\rotatebox{90}{\legendbox{violet} other flat}$ 
    & $\rotatebox{90}{\legendbox{teal} sidewalk}$ 
    & $\rotatebox{90}{\legendbox{olive} terrain}$ 
    & $\rotatebox{90}{\legendbox{lightgray} manmade}$ 
    & $\rotatebox{90}{\legendbox{green} vegetation}$ 
    \\
    \midrule
    BEVFormer \cite{li2022bevformer} &$256\times 704$ &29.2	&4.8	&35.0	&0.0	&35.5	&41.4	&10.4	&1.8	&17.5	&10.1	&23.3	&26.7	&79.3	&36.6	&47.7	&51.9	&39.6	&35.0\\
    SurroundOcc \cite{wei2023surroundocc} &$256\times 704$ &31.3	&9.2	&33.7	&17.5	&34.6	&39.9	&16.7	&18.4	&22.4	&20.8	&25.6	&27.3	&75.4	&35.1	&43.2	&47.1	&34.5	&30.2\\
    OpenOccupancy \cite{wang2023openoccupancy} &$256\times 704$ &32.5 &9.7 &40.2 &18.8 & 36.6 & 44.1 &8.1 &19.5 &22.9 &24.2 & 24.5 &28.9 &77.6 &37.3 &45.2 &49.2 &35.5  &30.6\\
    BEVDet-depth \cite{huang2021bevdet} &$256\times 704$ &31.4	&5.9	&38.7	&0.5	&38.6	&44.9	&14.4	&8.4	&17.0	&14.6	&24.1	&29.5	&79.7	&39.5	&49.6	&53.1	&39.8	&34.8\\
    BEVDet-stereo \cite{huang2021bevdet} &$256\times 704$ &34.8	&8.1	&42.0	&5.7	&41.3	&47.7	&21.0	&15.5	&19.4	&17.7	&29.8	&34.7	&\textbf{80.4}	&38.7	&\textbf{51.8}	&\textbf{55.1}	&\textbf{44.4}	&\textbf{38.8} \\
    \rowcolor{gray!20} 
    UniVision  &$256\times 704$ &\textbf{37.5} &\textbf{11.0} &\textbf{44.7} &\textbf{23.1} &\textbf{43.0} &\textbf{50.5} &\textbf{21.6} &\textbf{24.9} &\textbf{26.9} &\textbf{25.7} &\textbf{30.7} &\textbf{35.8} & 79.8 &\textbf{41.4} &49.1 &53.8 &40.3 &34.7\\
    \midrule
    BEVFormer \cite{li2022bevformer} &$512\times 1408$ &34.7	&7.1	&40.7	&9.8	&40.1	&47.8	&16.3	&17.5	&23.3	&20.7	&27.4	&33.2	&81.3	&39.7	&50.4	&54.3	&43.2	&36.8 \\
    SurroundOcc \cite{wei2023surroundocc} &$512\times 1408$ &34.4	&8.7	&39.2	&19.7	&41.4	&46.2	&18.7	&20.6	&26.4	&23.3	&27.0	&32.5	&78.0	&38.3	&46.6	&49.6	&36.7	&31.6 \\
    OpenOccupancy \cite{wang2023openoccupancy} &$512\times 1408$ &36.1 &10.4 &45.7 &23.6 &42.4 &49.3 &14.8 &24.6 &27.7 &27.8 &27.6 &33.3 &79.2 &39.8 &47.1 &50.5 &37.7 &31.8\\
    BEVDet-depth \cite{huang2021bevdet} &$512\times 1408$ &33.7	&6.6	&41.2	&7.0	&42.7	&48.4	&18.4	&12.9	&22.0	&18.2	&28.2	&33.2	&80.1	&39.7	&49.1	&52.1	&39.9	&33.8 \\
    BEVDet-stereo \cite{huang2021bevdet} &$512\times 1408$ &38.0	&8.6	&45.9	&14.3	&\textbf{46.0}	&51.2	&\textbf{23.8}	&18.9	&24.1	&22.3	&\textbf{33.6}	&37.9	&\textbf{81.5}	&40.5	&\textbf{52.6}	&\textbf{55.9}	&\textbf{46.9}	&\textbf{41.2} \\
    \rowcolor{gray!20} 
    UniVision  &$512\times 1408$ &\textbf{39.8}	&\textbf{11.3}	&\textbf{47.1}	&\textbf{27.6}	&45.8	&\textbf{54.2}	&22.9	&\textbf{28.6}	&\textbf{31.0}	&\textbf{28.7}	&31.8	&\textbf{38.5}	&81.2	&\textbf{42.5}	&51.3	&54.7	&41.7	&36.6 \\
    \bottomrule
    \end{tabular}}}
    \caption{Results on Occ3D benchmark. UniVision achieves state-of-the-art performance with different input image resolutions. 
    The results of the compared methods are reproduced with their official code base. 
    * Note that BEVFormer uses video input and BEVDet-stereo uses depth pre-training and stereo input but UniVision does not.
    }
    \label{tab:benchmark-occ-3d}
\end{table*}

\textbf{NuScenes 3D Object Detection.}
For the detection task, we use the official metrics of nuScenes, the nuScenes Detection Score (NDS), which is a weighted sum of mean Average Precision (mAP) and several metrics, including Average Translation Error (ATE), Average Scale Error (ASE), Average Orientation Error (AOE), Average Velocity Error (AVE) and Average Attribute Error (AAE).

\textbf{OpenOccupancy.} 
The OpenOccupancy benchmark \cite{wang2023openoccupancy} is based on the nuScenes dataset and provides semantic occupancy labels of $512 \times 512 \times 40$ resolution. The labeled classes are the same as those in the LiDAR segmentation task and we use mIoU as the evaluation metric.

\textbf{Occ3D.} 
The Occ3D benchmark \cite{tian2023occ3d} is based on the nuScenes dataset and provides semantic occupancy labels of $200 \times 200 \times 16$ resolution. 
Occ3D further provides visible masks for training and evaluation.
The labeled classes are the same as those in the LiDAR segmentation task and we use mIoU as the evaluation metric. 

\subsection{Experimental Settings}

\textbf{NuScenes LiDAR Segmentation}
For the LiDAR segmentation task, we use the sparse LiDAR segmentation labels as the supervision only and no extra dense occupancy labels from other benchmarks are used.
We use ResNet-101 \cite{resnet} as the image backbone and the image resolution is set to $896 \times 1600$.
The model is trained for 20 epochs with a total batch size of 32.
We use the AdamW \cite{loshchilov2017decoupledadamw} optimizer and the learning rate is set to 0.0002.
No temporal information or test time augmentation (TTA) is used.

\textbf{NuScenes Detection}
For the results on the nuScenes detection benchmark \cite{nuscenes}, we provide two versions of comparison results.
In the first version, we select three previous best methods \cite{li2022bevformer, huang2021bevdet, li2022bevdepth} and align the training settings including image backbone, input resolution, batch size, and learning rate to make a fair comparison.
We use the ResNet-50 image backbone and the models are trained for 20 epochs with a learning rate of 0.0002.
The batch size is set to 32 when the input resolution is $256 \times 704$ and 16 when the input resolution is $512 \times 1408$.

In the second version, we upscale the model and compare the results with those reported in other methods' papers.
We use ResNet-101 as the image backbone and the image resolution is set to $512 \times 1408$.
The model is trained for 20 epochs with a total batch size of 32 using the AdamW optimizer and the learning rate is set to 0.0002.
For UniVision4D, we initialize the model weights from the single-frame version and train the model for 10 epochs with a learning rate of 0.0001.
Notably, we do not use the CBGS \cite{zhu2019classcbgs} strategy in other methods.

\textbf{OpenOccupancy}
We use the ResNet-50 image backbone and the models are trained for 20 epochs with a learning rate of 0.0002, and the batch size is set to 32.
The input resolution is set to $512 \times 1408$.
Considering that the OpenOccupancy benchmark provides labels with a resolution of $512 \times 512 \times 40$ which is very memory-consuming, we down-sample the labels to half the original resolution for training.
During the inference phase, we up-sample the output to the original resolution.
No temporal information or TTA is used.

\textbf{Occ3D}
Considering that the Occ3D benchmark \cite{tian2023occ3d} is newly released with few reported results, we use the official codes of the compared methods \cite{li2022bevformer, wei2023surroundocc, wang2023openoccupancy, huang2021bevdet, li2022bevdepth} and align the training settings including image backbone, input resolution, batch size, and learning rate for a fair comparison.
We choose the ResNet-50 backbone and train the models for 20 epochs.
The batch size is set to 32 when the input resolution is $256 \times 704$ and 16 when the input resolution is $512 \times 1408$.
We use the AdamW optimizer and the learning rate is set to 0.0002.
During training and inference, the camera visible mask is used for loss calculation or metric evaluation.
No temporal information or TTA is used.

\begin{table*}[ht]
    \renewcommand{\arraystretch}{0.8}
    \footnotesize
    \centering
    \begin{tabular}{ccc|cc|ccccc}
    \toprule
    Global Branch & Occ $Sup.$ & Voxel $Inter.$ &mAP$\uparrow$ &NDS$\uparrow$ &mATE$\downarrow$ &mASE$\downarrow$ &mAOE$\downarrow$ &mAVE$\downarrow$ &mAAE$\downarrow$ \\
    \midrule
    $\times$    &$\times$   &$\times$       &0.271 &0.352 &0.725 &0.287 &0.663 &0.911 &0.256 \\
    \checkmark  &$\times$   &$\times$       &0.288 &0.382 &0.718 &0.291 &0.623 &\textbf{0.776} &0.214 \\
    \checkmark  &\checkmark &$\times$       &0.304 &0.384 &0.707 &0.285 &0.680 &0.789 &0.222 \\
    \checkmark  &$\times$   &\checkmark     &0.297 &0.389 &0.684 &0.285 &0.615 &0.804 &\textbf{0.197} \\
    \checkmark  &\checkmark &\checkmark     &\textbf{0.306} &\textbf{0.394} &\textbf{0.662} &\textbf{0.284} &\textbf{0.601} &0.840 &0.205 \\
    \bottomrule
    \end{tabular}
    \caption{
    Ablation study for the detection task. 
    Occ $Sup.$ denotes adding the voxel branch and using the occupancy task as auxiliary supervision without interaction. 
    Voxel $Inter.$ denotes explicitly using the voxel features for cross-representation interaction.}
    \label{tab:abl-det}
\end{table*}

\textbf{Abalation Studies}
For the ablation studies, we use the nuScenes detection benchmark \cite{nuscenes} and the Occ3D benchmark \cite{tian2023occ3d} to validate the effectiveness of the components.
We use the ResNet-50 backbone and the image resolution is set to $256 \times 704$.
All the models are trained for 20 epochs with a total batch size of 32.
We use the AdamW optimizer and the learning rate is set to 0.0002.

\subsection{Results}

\textbf{NuScenes LiDAR Segmentation.}
We show the results on nuScenes LiDAR Segmentation benchmark in \cref{tab:benchmark-nus-lidarseg}. 
UniVision significantly surpasses state-of-the-art (SOTA) vision-based method OccFormer \cite{zhang2023occformer} by 1.5 mIoU and sets a new record among vision-based models on the leaderboard. 
Notably, UniVision also outperforms some LiDAR-based models, such as PolarNet \cite{zhang2020polarnet} and DB-UNet \cite{wang2023dbunet}.

\textbf{NuScenes 3D Object Detection.}
As shown in \cref{tab:benchmark-nus-det}, when we use the same training settings for a fair comparison, UniVision shows superiority over other methods \cite{li2022bevformer, huang2021bevdet, li2022bevdepth}.
Specifically, UniVision achieves 2.4 points gain in mAP and 1.1 points gain in NDS against BEVDepth with the $512\times1408$ image resolution.
When we scale up the model and incorporate UniVision with temporal inputs, it further outperforms SOTA temporal-based detectors by a remarkable margin. UniVision achieves this with a smaller input resolution and it does not use the CBGS \cite{zhu2019classcbgs}.

\textbf{OpenOccupancy.} 
The results on the OpenOccupancy benchmark are shown in \cref{tab:benchmark-nus-openoccupancy}.
UniVision significantly surpasses recent vision-based occupancy methods including MonoScene \cite{Cao2021monoscene}, TPVFormer \cite{huang2023tpvformer} and C-CONet \cite{wang2023openoccupancy} by 7.3 points, 6.5 points and 1.5 points in mIoU, respectively.
Also, UniVision surpasses some LiDAR-based methods like LMSCNet \cite{roldao2020lmscnet} and JS3C-Net \cite{yan2021js3cnet}.

\textbf{Occ3D.} 
\cref{tab:benchmark-occ-3d} lists the results on the Occ3D benchmark.
With different input image resolutions, UniVision significantly outperforms recent vision-based methods \cite{li2022bevformer, wei2023surroundocc, huang2021bevdet} by more than 2.7 points and 1.8 points in mIoU.
Notably, BEVFormer and BEVDet-stereo load pre-trained weights and use temporal input in inference, while UniVision does not use them but still achieves better performance.

\subsection{Ablation Studies}

\textbf{Effectiveness of components in detection task.}
We show the ablation study for the detection task in \cref{tab:abl-det}.
When we insert the BEV-based global feature extraction branch into the baseline model, the performance is boosted by 1.7 mAP and 3.0 NDS.
When we add voxel-based occupancy task as an auxiliary task to the detector, the model has an improvement of 1.6 points gain in mAP.
When we explicitly introduce the cross-representation interaction from the voxel features, the model achieves the best performance, which is improved by 3.5 points in mAP and 4.2 points in NDS compared with the baseline.

\begin{table}
    \setlength{\tabcolsep}{3pt}
    \footnotesize
    \resizebox{\linewidth}{!}
    {
    \begin{tabular}{ccc|ccc}
    \toprule
    Local Branch & Det $Sup.$ & BEV $Inter.$  & mIoU & $\rm{mIoU}_{fore}$ &$\rm{mIoU}_{back}$\\
    \midrule
    $\times$    &$\times$   &$\times$       &34.58 &29.42 & 47.13  \\
    \checkmark  &$\times$   &$\times$       &36.52 &31.15 & 49.66  \\
    \checkmark  &\checkmark &$\times$       &36.96 &32.22 & 49.18   \\
    \checkmark  &$\times$   &\checkmark     &36.51 &30.93 & \textbf{49.99}  \\
    \checkmark  &\checkmark &\checkmark     &\textbf{37.12} &\textbf{32.37} &49.38 \\
    \bottomrule
    \end{tabular}
    }
    \caption{
    Ablation study for the occupancy task. 
    Det $Sup.$ denotes adding the BEV branch and using the detection task as auxiliary supervision without interaction. 
    BEV $Inter.$ denotes explicitly using the BEV features for cross-representation interaction.
    $\rm{mIoU}_{fore}$ denotes the mIoU of the foreground object classes.
    $\rm{mIoU}_{back}$ denotes the mIoU of the background object classes.
    }
    \label{tab:abl-occ}
\end{table}

\begin{table}
    \footnotesize
    \setlength{\tabcolsep}{3pt}
    \resizebox{\linewidth}{!}
    {
    \begin{tabular}{ccc|ccc}
    \toprule
    Occ-Det $Aug.$ & Ex-Im $Trans.$ & $Prog.$ $Adj.$  & mIoU & mAP &NDS\\
    \midrule
    $\times$    &-              &-                      &36.44 &0.271 & 0.372  \\
    \checkmark  &-              &-                      &\textbf{37.12} &\textbf{0.306} &\textbf{0.394}\\
    \midrule
    -           &$\times$       &-                      &36.96 &0.304 & 0.384  \\
    -           &\checkmark     &-                      &\textbf{37.46} &\textbf{0.312} &\textbf{0.397}  \\
    \midrule
    -           &-              &$\times$               & -- & -- & --   \\
    -           &-              &\checkmark             &\textbf{37.12} &\textbf{0.306} &\textbf{0.394}\\
    \bottomrule
    \end{tabular}
    }
    \caption{
    Ablation study for joint Occ-Det augmentation, Ex-Im view transform and progressive loss weight adjustment. -- denotes that the model can not converge and the performance is very low.}
    \label{tab:abl-other}
\end{table}

\textbf{Effectiveness of components in occupancy task.}
We show the ablation study for the occupancy task in \cref{tab:abl-occ}.
The voxel-based local feature extraction network brings an improvement of 1.96 mIoU gain to the baseline model.
When the detection task is introduced as an auxiliary supervision signal, the model performance is boosted by $0.4$ in mIoU.
When we explicitly fuse the glocal BEV features with the voxel features, the model achieves the best performance with 2.64 points gain in mIoU compared with the baseline.

\begin{figure*}
    \centering
    \includegraphics[width=1\textwidth]{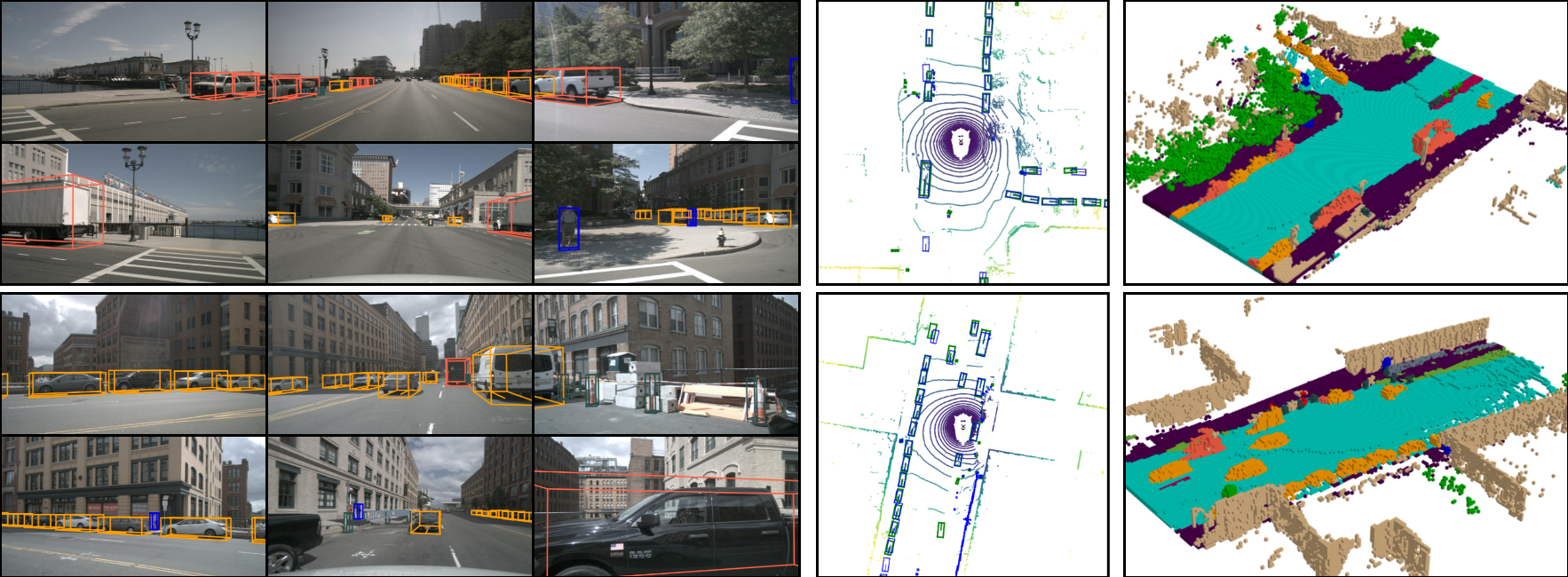}
    \caption{Qualitative results of UniVision framework, including the detection results on the 2D image plane, the detection results on the BEV plane, and the corresponding occupancy prediction results.}
    \label{fig:qualatative results}
\end{figure*}

\textbf{How do detection task and occupancy task influence each other?}
\cref{tab:abl-det} and \cref{tab:abl-occ} show that both the detection task and the occupancy task benefit each other in our UniVision framework.
For the detection task, the occupancy supervision can improve the mAP and mATE metrics, which indicates that voxel-wise semantic learning effectively improves the detector's awareness of object geometry, \ie, centerness and scale.
For the occupancy task, the detection supervision significantly improves the performance for foreground categories, \ie, detection categories, thus resulting in an overall improvement.

\textbf{Effectiveness of joint Occ-Det augmentation, Ex-Im view transform and progressive loss weight adjustment.}
We show the effectiveness of the joint Occ-Det spatial augmentation, the Ex-Im view transform module, and the progressive loss weight adjustment strategy in \cref{tab:abl-other}.
It shows significant improvements in the detection task and the occupancy task with the proposed spatial augmentation and the proposed view transform module on the mIoU, mAP, and NDS metrics.
The loss weight adjustment strategy enables the efficient training of the multi-task framework. Without this, the training of the unified framework cannot converge and the performance is very low.

\subsection{Qualitative Results}
We show qualitative results of UniVision in \cref{fig:qualatative results}, which include the detection results on the 2D image plane, the detection results on the BEV plane, and the corresponding occupancy prediction results.
The UniVision framework can simultaneously produce high-quality prediction results for both 3D detection and occupancy prediction tasks with a unified network.

\section{Conclusion}
In this paper, we present UniVision, a unified framework for vision-centric 3D perception that can simultaneously handle the occupancy prediction task and the 3D detection task.
To achieve this, we propose the explicit-implicit view transform module for complementary 2D-3D feature transformation.
We propose a local-global feature extraction and fusion module for efficient and adaptive multi-representation feature extraction, enhancement, and interaction.
Besides, we propose a joint occ-det data augmentation strategy and a progressive loss weight adjustment strategy for efficient and stable multi-task framework training.
UniVision achieves state-of-the-art performance on four benchmarks, including nuScenes LiDAR segmentation, nuScenes detection, OpenOccupancy, and Occ3D.

\clearpage

{\small
\bibliographystyle{ieee_fullname}
\bibliography{PaperForReview}
}

\end{document}